\documentclass[10pt,twocolumn,letterpaper]{article}

\usepackage[pagenumbers]{cvpr} 
\usepackage{times,latexsym,multirow}
\usepackage[skip=3pt,font=small]{subcaption}
\usepackage[skip=3pt,font=small]{caption}
\usepackage[table]{xcolor}
\usepackage{acronym}
\usepackage{amsfonts}
\usepackage[misc]{ifsym}
\definecolor{cvprblue}{rgb}{0.21,0.49,0.74}
\usepackage[pagebackref,breaklinks,colorlinks,allcolors=cvprblue]{hyperref}

\makeatletter
\DeclareRobustCommand\onedot{\futurelet\@let@token\@onedot}
\def\@onedot{\ifx\@let@token.\else.\null\fi\xspace}
\def\eg{\emph{e.g}\onedot}

\makeatother

\makeatletter
\renewcommand{\paragraph}{%
  \@startsection{paragraph}{4}%
  {\z@}{0ex \@plus 0ex \@minus 0ex}{-1em}%
  {\hskip\parindent\normalfont\normalsize\bfseries}%
}
\makeatother

\newcommand{\system}{\textbf{EveryWear}\xspace}
\newcommand{\dataset}{\textbf{Ego-Elec}\xspace}

\newcommand{\rmnum}[1]{\romannumeral #1}

\acrodef{mocap}[MoCap]{motion capture}
\acrodef{mpjpe}[MPJPE]{Mean Per Joint Position Error}
\acrodef{pa-mpjpe}[PA-MPJPE]{Procrustes Aligned Mean Joint Position Error}
\acrodef{mpjre}[MPJRE]{Mean Per Joint Rotation Error}
\acrodef{mpjve}[MPJVE]{Mean Per Joint Vertex Error}
\acrodef{root-pe}[Root PE]{Root Position Error}

\newcommand{\supp}{\textit{supplementary material}\xspace}

\title{Human Motion Estimation with Everyday Wearables}

\author{
        Siqi Zhu\textsuperscript{1}\footnotemark[1]\quad 
        Yixuan Li\textsuperscript{1}\footnotemark[1]\quad
        Junfu Li\textsuperscript{1,2}\footnotemark[1]\quad
        Qi Wu\textsuperscript{1,2}\footnotemark[1]\\
        Zan Wang\textsuperscript{1}\footnotemark[1]\quad
        Haozhe Ma\textsuperscript{3}\quad
        Wei Liang\textsuperscript{1,2}$^{\,\textrm{\Letter}}$\\
        \small \footnotemark[1]\;\;Equal contributors \quad
        \textsuperscript{1} Beijing Institute of Technology\quad\\
        \small\textsuperscript{2} Yangtze Delta Region Academy of Beijing Institute of Technology, Jiaxing\quad
        \textsuperscript{3} Shenzhen MSU-BIT University\vspace{6pt}\\
        \href{https://pie-lab.cn/EveryWear/}{https://pie-lab.cn/EveryWear/}
        \vspace{-15pt}
}

\begin{document}

\twocolumn[{%
\renewcommand\twocolumn[1][]{#1}%
\maketitle
\begin{center}
    \captionsetup{type=figure}
    \includegraphics[width=\linewidth]{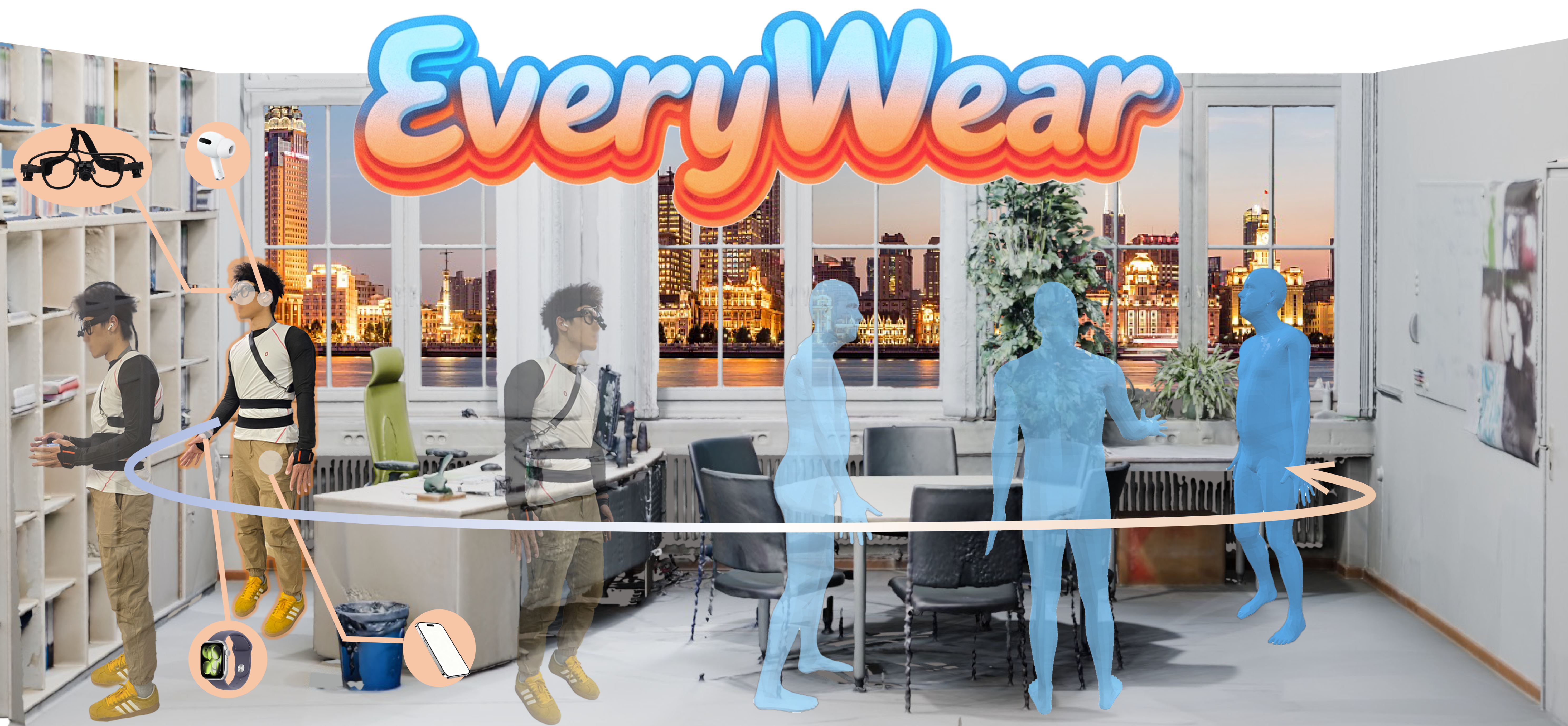}
    \captionof{figure}{\system is a novel human motion capture approach based on a series of lightweight everyday wearables: a smartphone, smartwatch, earbuds, and smart glasses equipped with one forward-facing and two downward-facing cameras.}
    \label{fig:teaser}
\end{center}
}]

\begin{abstract}
While on-body device-based human motion estimation is crucial for applications such as XR interaction, existing methods often suffer from poor wearability, expensive hardware, and cumbersome calibration, which hinder their adoption in daily life. 
To address these challenges, we present \system, a lightweight and practical human motion capture approach based entirely on everyday wearables: a smartphone, smartwatch, earbuds, and smart glasses equipped with one forward-facing and two downward-facing cameras, requiring no explicit calibration before use.
We introduce \dataset, a $9$-hour real-world dataset covering $56$ daily activities across $17$ diverse indoor and outdoor environments, with ground-truth 3D annotations provided by the \acf{mocap}, to facilitate robust research and benchmarking in this direction.
Our approach employs a multimodal teacher-student framework that integrates visual cues from egocentric cameras with inertial signals from consumer devices. By training directly on real-world data rather than synthetic data, our model effectively eliminates the sim-to-real gap that constrains prior work.
Experiments demonstrate that our method outperforms baseline models, validating its effectiveness for practical full-body motion estimation.
\end{abstract}

\section{Introduction}
Human motion estimation is crucial for natural interaction and realistic avatar animation in XR applications~\citep{xu2024mobileposer,ym2025physic,xie2025intertrack,zhang2024force,hollidt2024egosim,jiang2024egoposer}, as well as expressive humanoid whole-body control~\citep{ji2024exbody2,li2025clone,yin2025visualmimic, du2025learning}. 
Prior work on estimating human motion follows two main directions. Camera-based methods~\citep{rhodin2016egocap,wang2023scene,akada20243d,wang2021estimating,xu2019mo,camiletto2025frame,tome2019xr,akada2022unrealego,zhao2021egoglass} suffer from self-occlusion and environmental occlusions, limiting real-world robustness. IMU-based approaches~\citep{mollyn2023imuposer,xu2024mobileposer,yi2021transpose,huang2018deep} mitigate occlusion issues but struggle with cumulative drift and sensor instability, degrading long-term accuracy.

To address these limitations, recent work ~\citep{fan2025emhi,lee2024mocap,yi2023egolocate,guzov2021human,Cha2021Mobile} has explored fusing visual and inertial information to improve accuracy.  Although these approaches improve accuracy, several practical factors still limit their deployment: (\rmnum{1}) \textbf{Limited Wearability}, as they often rely on bulky head-mounted cameras or dense IMU configurations; (\rmnum{2}) \textbf{Non-trivial Calibration}, requiring careful alignment across multiple heterogeneous sensors; and (\rmnum{3}) \textbf{High Hardware Cost}, since professional-grade IMUs remain expensive and inaccessible. 
To this end, we address these issues by \textbf{enabling human motion estimation using only lightweight, calibration-free everyday wearables}.

In this paper, we propose a novel human motion estimation framework, \system, that leverages RGB images from glasses with three cameras (one forward, two downward) and inertial data from typical wearables: a smartphone, a smartwatch, and earbuds, all carried in everyday configurations. This setup is lightweight and practical for daily use.
However, these sparse consumer-grade sensors also introduce unique challenges for accurate motion estimation: 
(\rmnum{1}) \textbf{Low sensor stability.} Consumer IMUs are inherently less stable than professional hardware, due to lower sensor quality (data loss, signal noise) and loose everyday wear configurations (phones in pockets, watches with play). The complex instability makes realistic sensor modeling difficult, leading to the failure of previous synthetic-data-based training approaches built on idealized assumptions.
(\rmnum{2}) \textbf{Limited motion observability.} The sparse sensors setup provides only partial body motion observations compared to dense professional IMU configurations, especially in camera occluded scenarios, leading to incomplete constraints for accurate full-body motion estimation.


To address these challenges, we propose a teacher-student distillation approach. The teacher policy leverages multimodal inputs: RGB images from three cameras and IMU data from two wrists, two legs, and the head obtained from \ac{mocap} system~\citep{xsens}, to learn accurate motion estimation. We then distill this knowledge into a student policy that uses only sparse, noisy IMU measurements from consumer devices while maintaining the same camera inputs. This distillation process constrains the student to learn from the teacher's knowledge obtained from precise sensor observations, while simultaneously adapting to noisy consumer-grade IMU data. 

Importantly, while sparse sensors provide incomplete motion constraints individually, the multimodal design enables cross-modal compensation: when one modality becomes unreliable (\eg, cameras are occluded or IMU drift occurs), other modalities compensate to maintain robust motion estimation.
Furthermore, we integrate an off-the-shelf SLAM module using the forward-facing camera to provide global localization and compensate for drift in both position and head pose estimation.

To facilitate further research in this direction, we introduce \dataset, a large-scale dataset comprising RGB images from three cameras and IMU data from three consumer devices collected using the prototype of \system. The dataset features $9$ hours of real-world human motion across $56$ types of daily activities in $17$ diverse environments, with ground-truth $3$D body poses and global translations annotated using the
\acf{mocap} system. 
By training directly on real-world data, our approach achieves robust generalization to real-world scenarios without the sim-to-real gap that limits synthetic-data-based methods. Moreover, the diverse activities and environments make our dataset valuable for future research and broadly applicable across various applications.

We conduct comprehensive experiments to validate the effectiveness of our approach. Our method achieves $8.459$ cm MPJPE and $10.627$ cm MPJVE, significantly outperforming baselines by $3.345$ cm and $4.289$ cm, demonstrating that distillation from accurate teacher observations helps the model adapt to noisy consumer-grade sensors. Ablation studies further validate our design choices, confirming that (\rmnum{1}) multimodal fusion enables robust cross-modal compensation, and (\rmnum{2}) our approach maintains performance under occlusion scenarios where single-modality methods fail.

In summary, our contributions are as follows:
\begin{itemize}[leftmargin=*,nolistsep,noitemsep]
    \item A practical motion capture framework using only everyday consumer devices~(smartphone, smartwatch, earbuds, smart glasses) with no calibration required, achieving robust motion estimation through teacher-student distillation and multimodal fusion.
    \item Ego-Elec, a comprehensive real-world dataset with $9$ hours of egocentric motion across $56$ activities and $17$ environments, the first large-scale dataset combining egocentric vision with sparse consumer IMU measurements.
    \item State-of-the-art results demonstrating that our approach outperforms existing methods while using only consumer devices, with robust performance maintained under occlusions through cross-modal compensation.
\end{itemize}

\section{Related Work}
\subsection{On-Body Device-Based Motion Capture}
On-body device-based human motion capture systems fall into three categories: IMU-only, camera-only, and multimodal approaches. 
\textbf{IMU-only methods} use sparse sensors attached to body locations such as wrists, knees, or other joints~\citep{huang2018deep,xiao2024fast,van2024diffusionposer,zhu2025progressive}. Early work used professional IMUs, which are costly and inaccessible for daily use. Recent approaches~\citep{mollyn2023imuposer,xu2024mobileposer,devrio2023smartposer} leverage consumer electronics~(smartphones, smartwatches), but sparse consumer IMU configurations provide limited body coverage, constraining motion estimation accuracy. 
\textbf{Camera-only methods} employ egocentric cameras mounted on the head~\citep{rhodin2016egocap,wang2022egopw,akada2025ego4view,tome2019xr,zhao2021egoglass} or chest~\citep{nishikawa2025egocentric}. While cameras provide rich visual observations, they struggle with occlusions caused by furniture, walls, or body parts blocking the camera's view.
\textbf{Multimodal methods} combine head-mounted cameras, chest cameras, and inertial sensors for comprehensive observations and cross-modal compensation~\citep{grauman2024egoexo4d,wang2023holoassist, fan2025emhi,ma2024nymeria}. However, these systems burden users with multiple body-mounted sensors and 
complex calibration procedures, limiting practical deployment. 
In contrast, we adopt a lightweight configuration centered on glasses-mounted cameras, complemented by everyday consumer devices. This design preserves comprehensive egocentric observations through multimodal fusion while avoiding the hardware burden and calibration requirements of existing systems, making it practical for daily use.

\subsection{On-Body-Device-Based Motion Dataset}

Corresponding to the three types of motion capture methods discussed above, existing datasets for human motion estimation can also be categorized in the same way: camera-only~\citep{akada2025ego4view,akada2022unrealego}, IMU-only~\citep{mahmood2019amass,mollyn2023imuposer,devrio2023smartposer,xu2024mobileposer}, and multimodal approaches~\citep{grauman2024egoexo4d,grauman2022ego4d,liu2022hoi4d,wang2023holoassist,akada2022unrealego,yang2025egolife,guzov2021hps}. 
These datasets have significantly advanced human motion estimation research by providing diverse activities and motion patterns. However, most of these works rely on synthetic data to avoid the labor-intensive process of collecting and annotating real-world data.
While synthetic data enables easy scaling with different sensor modalities, it introduces a significant sim-to-real gap that limits real-world deployment. First, real-world IMU measurements are inherently noisy and affected by environmental disturbances (magnetic interference, temperature drift) that are difficult to accurately simulate. Second, everyday wear configurations, such as phones loosely placed in pockets or watches with some wrist play, introduce relative movement between sensors and the body that is challenging to model in simulation. Recent efforts~\citep{akada2025ego4view,hollidt2024egosim,xu2019mo,liu2022hoi4d,wang2023holoassist,sun2025suite} have attempted to combine synthetic and real data, but they still inherit the fundamental limitations of synthetic sensor modeling.
To address this gap, we introduce the first large-scale real-world dataset that combines egocentric vision with sparse consumer IMU measurements. By training directly on real-world data, our approach eliminates the sim-to-real gap and enables robust deployment in practical scenarios.

\subsection{Egocentric Motion Estimation}
Previous single-modality approaches~\citep{cuevas2024simpleego,shen2025fish2mesh,akada2025ego4view} process camera or IMU inputs independently, either through end-to-end regression or via $2$D heatmap as intermediate representations. Multimodal methods~\citep{gilbert2019fusing} process modalities separately and fuse pose estimate results directly, failing to exploit cross-modal feature interactions.
Critically, most existing methods train on synthetic IMU data~\citep{akada2022unrealego,akada2025ego4view} and train directly on sparse sensors, limiting their applicability to real-world scenarios with sparse, noisy consumer devices. Our teacher-student distillation framework addresses these limitations by (\rmnum{1}) training on real-world data to capture true noise characteristics, and (\rmnum{2}) learning to estimate poses from sparse consumer sensors through knowledge transfer from a teacher trained on dense, accurate sensors. Our multimodal fusion enables robust cross-modal compensation when individual sensors become unreliable.

\begin{figure}[t!]
    \centering
    \includegraphics[width=\linewidth]{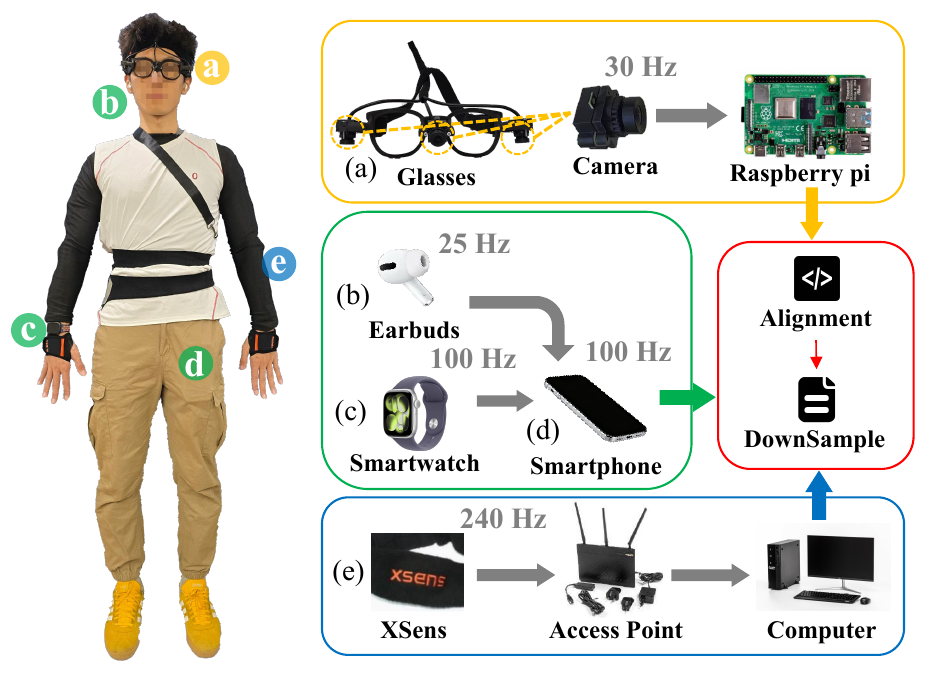}
    \caption{\textbf{System illustration.} The system comprises smart glasses (with onboard Raspberry Pi), smartphone, smartwatch, and earbuds. An XSens \ac{mocap} system provides annotations.}
    \label{fig:system}
\end{figure}

\section{Setup and Dataset}
\subsection{Hardware Setup}\label{sec:hardware}
As shown in \cref{fig:system}, our system comprises everyday consumer devices: a smartphone, a smartwatch, wireless earbuds, and custom smart glasses equipped with three cameras (one forward-facing, two downward-facing).
All cameras are connected to an onboard Raspberry Pi $5$, capturing synchronized RGB images at a resolution of $1080\times720$ and a frame rate of $30$ Hz. IMU data from the smartphone, smartwatch, and earbuds are recorded at $100$, $100$, and $25$ Hz, respectively.

During data collection, users were equipped with the system in typical everyday configurations: smart glasses on the head, a smartwatch on either wrist, earbuds in both ears, and a smartphone loosely placed in a pants pocket (left or right).
To obtain ground-truth annotations, participants simultaneously wear the \ac{mocap} system~\citep{xsens}, which provides $3$D joint positions and global translations at $240$ Hz using $17$ body-mounted IMU sensors.

\subsection{Multimodality Data Alignment}
Since the cameras, IMU sensors, and motion capture system operate independently, their data streams are neither temporally synchronized nor uniformly sampled. To address this, participants perform predefined calibration gestures at the start and end of each recording session, which we use to temporally align all modalities to a standard reference frame (details in \supp). We then downsample all streams to $25$ Hz, the earbuds' native rate, and the lowest in our system, using nearest-neighbor sampling.

\begin{table}[t]
    \centering
    \caption{Comparison with existing on-body device motion estimation datasets. \textbf{Type}: synthetic or real-world; \textbf{I/O}: indoor (\textbf{I}) or outdoor (\textbf{O}); \textbf{Cams}: number of cameras; \textbf{IMUs}: number of IMUs; \textbf{Acts}: number of activity types; \textbf{Frames}: number of frames.}
    \label{table:comparison_dataset}
    \resizebox{\linewidth}{!}{%
        \begin{tabular}{lccccccccc}
            \hline
            Datasets & Type & I/O & Cams & IMUs & Acts & Frames \\
            \hline
            M2C2~\cite{xu2019mo} & Syn. & - & 1 & - & 3K & 530K \\
            xR-EP~\cite{tome2019xr} & Syn. & - & 1 & - & 9 & 380K \\
            UnrealEgo~\cite{akada2022unrealego} & Syn. & - & 2 & - & 30 & 900K \\xw
            EgoCap~\cite{rhodin2016egocap} & Real & I/O & 2 & - & - & 75K \\
            EgoGlass~\cite{zhao2021egoglass} & Real & I & 2 & - & 6 & 173K \\
            EgoPW~\cite{wang2022egopw} & Real & I/O & 1 & - & 20 & 318K \\
            DIP-IMU~\cite{huang2018deep} & Real & I & - & 6 & 15 & 330K \\
            IMUPoser~\cite{mollyn2023imuposer} & Real & I & - & 3 & 36 & - \\
            FRAME~\cite{camiletto2025frame} & Real & I & 2 & - & 50 & 1.6M \\
            MEV-R~\cite{hollidt2024egosim} & Real & O & 6 & - & 35 & 3.12M \\
            Ego4View-R~\cite{akada2025ego4view} & Real & I & 4 & - & - & 930K \\
            ColossusEgo~\cite{lee2025rewind} & Real & I & 2 & - & - & 2.8M \\
            Ego-VIP~\cite{Cha2021Mobile} & Real & I & 4 & 4 & - & 38K \\
            EMHI~\cite{fan2025emhi} & Real & I & 2 & 5 & 39 & 3.07M \\
            \hline
            \bf Ego-Elec & Real & I/O & 3 & 3 & 56 & 2.88M \\
            \hline
        \end{tabular}
    }%
\end{table}

\subsection{\dataset Dataset}
Our dataset was collected using the hardware setup described in~\cref{sec:hardware}. Totally, our dataset provides \textbf{$9$ hours} of real-world human motion across \textbf{$56$ types} of daily activities in \textbf{$17$ diverse environments}. To contextualize its contributions, we compare it with existing datasets in~\cref{table:comparison_dataset}. As shown, we provide the most diverse and comprehensive real-world data across modalities and environments, comparable in scale to prior work.
As visualized in \supp, each recording session consists of $5$-$10$ scripted daily activities performed continuously in one of $17$ distinct indoor and outdoor environments, with an average duration of $4$ minutes per session. Activities are drawn from a taxonomy of $56$ distinct daily activity types~(details in \supp), including sports, social interactions, household tasks, and work-related actions. This design captures naturalistic motion sequences suitable for both motion estimation and long-horizon egocentric tasks.

\begin{figure*}[t!]
    \centering
    \includegraphics[width=\linewidth]{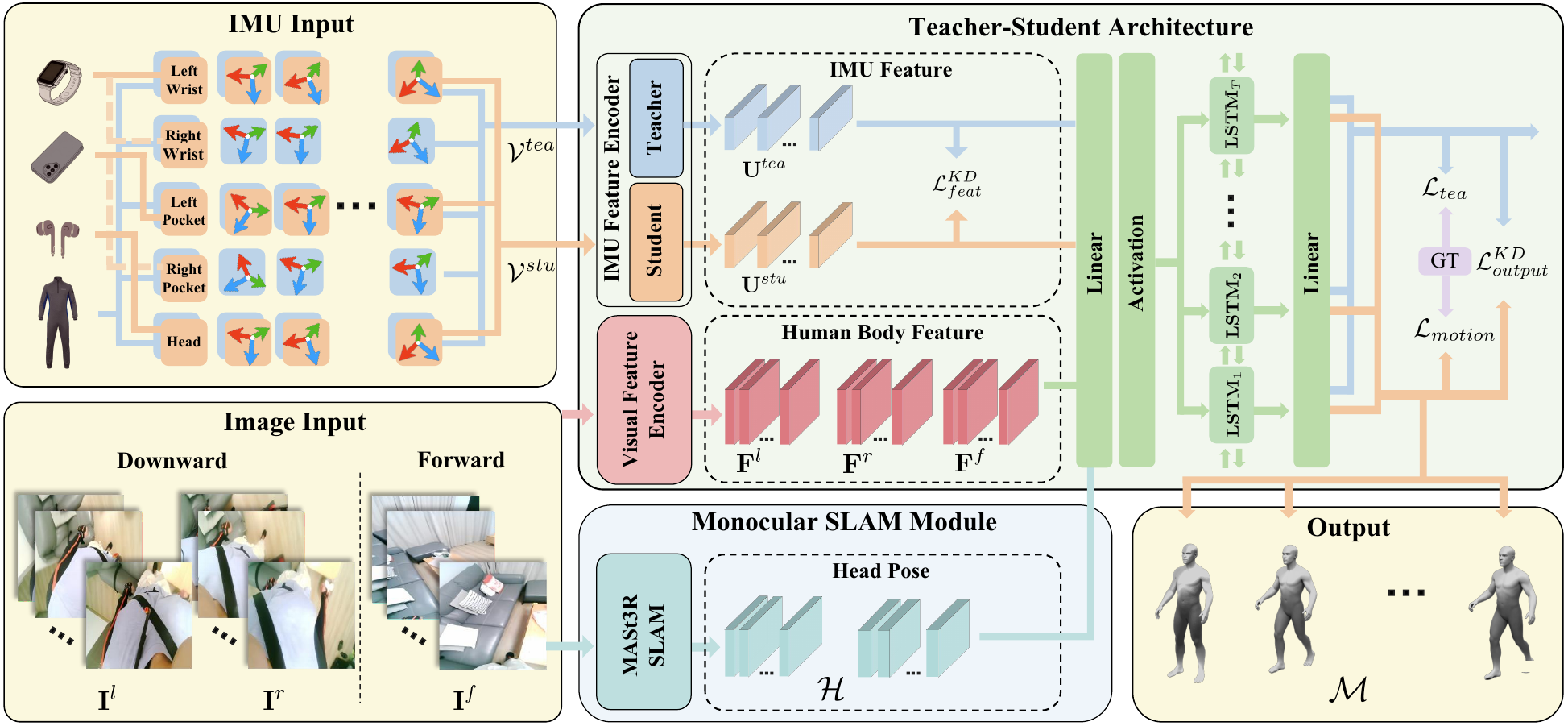}
    \caption{\textbf{Pipeline.} Our method takes egocentric images (three cameras) and IMU signals (everyday wearables) as input. We employ SLAM (\cref{sec:slam}) for head pose estimation, then use a teacher-student framework (\cref{sec:ts_model}) with shared visual feature encoder and separate IMU feature encoders, followed by bidirectional LSTM for temporal fusion and motion prediction.
}
    \label{fig:pipeline}
\end{figure*}

\section{Method}
\subsection{Overview}
Our objective is to estimate human motion $\mathcal{M}=\{\mathbf{P}_t, \mathbf{\Theta}_t\}_{t=0}^{T}$, where $\mathbf{P}_t \in \mathbb{R}^3$ is global root translation and $\mathbf{\Theta}_t=\{\theta_{j}\}_{j=0}^{J} \in \mathbb{R}^{24 \times 6}$ contains $J = 24$ joint rotations in $6$D representation~\citep{zhou2019continuity} for SMPL~\citep{loper2023smpl} at time $t$ over sequence length $T$.

We propose a teacher-student model (detailed in \cref{sec:ts_model}) that takes multi-view images $\mathcal{I} = \{\mathbf{I}^{f}_{t}, \mathbf{I}^{l}_{t}, \mathbf{I}^{r}_{t}\}_{t=0}^{T}$ and IMU signals $\mathcal{V}_{t=0}^{T}$ from five body-worn sensors as input to estimate motion sequence $\mathcal{M}$. Additionally, we employ an off-the-shelf SLAM module (\cref{sec:slam}) to estimate global head position, complementing the earbud's orientation measurements.
$\mathbf{I}^{f}_t, \mathbf{I}^{l}_t, \mathbf{I}^{r}_t$ denote frames from forward-facing, left downward-facing, and right downward-facing cameras at time $t$. IMU signals $\mathcal{V} \subseteq \{\mathbf{V}^{h}, \mathbf{V}^{lw}, \mathbf{V}^{rw}, \mathbf{V}^{lh}, \mathbf{V}^{rh}\}$ come from: head (earbuds), left wrist (smartwatch), right wrist (smartwatch), left hip (smartphone), and right hip (smartphone), respectively. Each sensor provides $\mathbf{V} = \{\mathbf{R}_{t}, \mathbf{A}_{t}\}_{t=0}^{T}$, where $\mathbf{R}_{t}$ and $\mathbf{A}_{t}$ represent orientation and acceleration respectively. The overall architecture of our model is illustrated in~\cref{fig:pipeline}.

\subsection{Monocular SLAM Module}
\label{sec:slam}
We employ the off-the-shelf SLAM module MASt3R-SLAM~\citep{murai2025mast3r} to estimate global camera poses $\mathbf{C} = \{\mathbf{C}_t\}_{t=0}^T$ from the forward camera sequence $\{\mathbf{I}^{f}_t\}_{t=0}^{T}$. We then apply a rigid transformation to obtain head poses $\mathcal{H} = \{\mathbf{H}_t\}_{t=0}^{T}$ from the camera poses $\mathbf{C}$.
Although the global translation scale differs between SLAM and IMU coordinate systems, our teacher-student model implicitly learns to align these frames during training.


\subsection{Teacher-Student Model}
\label{sec:ts_model}
\subsubsection{Teacher Model}
We formulate motion estimation as a sequence-to-sequence problem using a sliding window of length $N$. The model takes visual and IMU observations as input, extracts features using visual and IMU feature encoders, and then fuses the multimodal features through a temporal fusion module to estimate human motion, as illustrated in~\cref{fig:pipeline}.

\textbf{Visual Feature Encoder} We employ a ResNet-18 backbone pre-trained on ImageNet to extract visual features $\mathbf{F}_t^{f}$, $\mathbf{F}_t^{l}$, and $\mathbf{F}_t^{r}$ from input images $\mathbf{I}_t^{f}$, $\mathbf{I}_t^{l}$, $\mathbf{I}_t^{r}$ at each time step $t$.

\textbf{IMU Feature Encoder} An MLP-based encoder processes all IMU signals $\mathcal{V}^{tea}_t = \{\mathbf{V}_t^{h}, \mathbf{V}_t^{lw}, \mathbf{V}_t^{rw}, \mathbf{V}_t^{lh}, \mathbf{V}_t^{rh}\}$ into a unified latent representation $\mathbf{U}_t$. Note that during teacher training, we use dense IMU data from the motion capture system ($5$ IMUs) rather than the sparse consumer sensors ($3$ IMUs), enabling the teacher to learn from more accurate and complete motion 
observations.

\textbf{Temporal Fusion} A bidirectional LSTM processes the concatenated 
features $\{\mathbf{F}_t^{f}, \mathbf{F}_t^{l}, \mathbf{F}_t^{r}, \mathbf{U}_t, \mathbf{H}_t\}_{t}^{t+N}$ over the sliding window to estimate motion sequence $\mathbf{M}$, where $\mathbf{H}_t$ denotes the SLAM-derived head pose detailed in~\cref{sec:slam}.

 
 

\paragraph{Loss Function}
The teacher model is jointly optimized with two objectives: the local pose loss $\mathcal{L}_{pose}$ and the global translation loss $\mathcal{L}_{trans}$.
Following~\citet{kendall2018multi}, we use learned task-dependent uncertainty weighting. The total teacher loss is:
\begin{equation}
\label{loss:tea}
\mathcal{L}_{tea} = \frac{1}{\sigma_{pose}^2} \mathcal{L}_{pose} + \log \sigma_{pose}^2 + \frac{1}{\sigma_{trans}^2} \mathcal{L}_{trans} + \log \sigma_{trans}^2,
\end{equation}
where $\sigma_{pose}^2$ and $\sigma_{trans}^2$ are learnable 
uncertainty parameters that adaptively weight the losses during training.

\textbf{Local Pose Loss}  The pose loss is a weighted mean squared error over the $6$D rotation representations of all $J=24$ body joints:
\begin{equation}
\label{loss:local_pose}
\mathcal{L}_{pose} = \frac{1}{J}\sum_{j=1}^{J} w_j \,\big\| \mathbf{\theta}_j - \widehat{\mathbf{\theta}}_j \big\|_2^2,
\end{equation}
where $\mathbf{\theta}, \widehat{\mathbf{\theta}}\in\mathbb{R}^{6}$ 
are ground-truth and predicted rotations, and $w_j>0$ are per-joint weights 
(Refer to~\supp\ for detailed weights design.).

\textbf{Global Translation Loss} We implement the translation loss as the L$2$ norm between the predicted and ground-truth root positions: $\mathcal{L}_{trans} = \left\| \mathbf{P} - \widehat{\mathbf{P}} \right\|_2^2$,
where $\mathbf{P}, \widehat{\mathbf{P}}\in\mathbb{R}^3$ are ground-truth and 
predicted root positions.

\subsubsection{Student Model}
The student model shares the teacher's architecture but processes sparse IMU 
signals from consumer devices. While the teacher uses dense IMU data from $5$ 
body locations, the student uses only $3$ physical sensors: head (earbuds), wrist 
(smartwatch), and hips (smartphone): $\mathcal{V}_t^{stu} = 
\{\mathbf{V}_t^{head}, \mathbf{V}_t^{w}, \mathbf{V}_t^{hip}\}$. This reflects 
typical daily usage with one watch and one phone.

The student model's parameters are initialized with the teacher's pre-trained weights and fine-tuned with a small learning rate to adapt to noisy consumer IMU signals. Training details are provided in the \supp.

\paragraph{Loss Functions}
The student model is trained with a multi-objective loss that combines three components: (\rmnum{1}) a motion estimation loss~$\mathcal{L}_{motion}$, which is the same as the $\mathcal{L}_{tea}$. (\rmnum{2}) an distillation loss~$\mathcal{L}^{KD}_{output}$ that makes the student's output $\mathbf{M}=\{\mathbf{P}_t, \mathbf{\Theta}_t\}_{t=0}^{T}$ as same as the teacher’s. (\rmnum{3}) a feature-level distillation loss $\mathcal{L}_{feat}^{KD}$ that aligns the student’s IMU feature with those in the teacher model. 

The loss $\mathcal{L}_{output}^{KD}$ is defined as:
\begin{equation}
\mathcal{L}^{KD}_{output}
= \frac{1}{J}\sum_{j=1}^{J}
\left\| \widehat{\mathbf{\theta}}_{j}^{T} - \widehat{\mathbf{\theta}}_{j}^{S} \right\|_2^2,
\end{equation}
where $\widehat{\mathbf{\theta}}_{j}^{T}$ and $\widehat{\mathbf{\theta}}_{j}^{S}$ denote the predicted 6D joint rotations of joint $j$ from the teacher and student models, respectively.

The loss $\mathcal{L}_{feat}^{KD}$ is defined as: $\mathcal{L}_{feat}^{KD} = \frac{1}{d}\left\| \mathbf{U}^{tea} - \mathbf{U}^{stu} \right\|_2^2$,
where $\mathbf{U}^{tea}$ and $\mathbf{U}^{stu}$ represent the extracted IMU features in teacher and student models, respectively.

Formally, the total student loss $\mathcal{L}_{stu}$ is defined as:
\begin{equation}
\label{loss:stu}
\mathcal{L}_{stu}
= \lambda_{motion}\;\mathcal{L}_{motion}
+ \lambda_{output}\;\mathcal{L}^{KD}_{output}
+ \lambda_{feat}\;\mathcal{L}^{KD}_{feat},
\end{equation}
where $\lambda_*$ are the balanced weights. We provide detailed weight settings in the \supp.

\section{Experiments}

\begin{table*}[t]
    \centering
    \caption{Evaluation of \system on \dataset. The first section compares our model with baseline methods, while the second section presents ablation studies of our model design.}
    \label{table:comparison}
    \resizebox{\textwidth}{!}{%
        \begin{tabular}{llcccccc} 
            \toprule
             & Method & MPJPE (cm) & PA-MPJPE (cm) & MPJRE (deg) & MPJVE (cm) & Root PE (cm) \\
            \midrule
            \multirow{6}{*}{\textbf{Baseline}} 
            & \multicolumn{6}{>{\columncolor{lightgray!50}}l}{\textbf{(a) IMU-Only Method}} \\
            \cmidrule(l){2-7}
            & IMUPoser \cite{mollyn2023imuposer} & $11.804$ & $7.647$ & $12.300$ & $14.916$ & $24.499$ \\
            & MobilePoser  \cite{xu2024mobileposer} & $17.384$ & $15.099$ & $20.818$ & $17.303$ & $25.028$ \\
            \cmidrule(l){2-7}
            & \multicolumn{6}{>{\columncolor{lightgray!50}}l}{\textbf{(b) Camera-Only Method}} \\
            \cmidrule(l){2-7}
            & Fish2Mesh \cite{shen2025fish2mesh} & $12.677$ & $7.049$ & $11.036$ & $16.213$ & - \\
            \midrule
            \multirow{12.6}{*}{\textbf{Ablation}}
            & \multicolumn{6}{>{\columncolor{lightgray!50}}l}{\textbf{(a) Ablation on IMUs}} \\
            \cmidrule(l){2-7}
            & Ours-w/o-IMUs   & $13.014$  & $7.249$ & $10.901$ & $16.650$  & - \\
            & Ours-w/o-Phone  & $9.447$  & $6.183$ & $9.968$ & $11.786$ & $14.774$ \\
            & Ours-w/o-Watch  & $9.642$ & $6.253$ & $9.977$ & $12.121$  & $13.835$ \\
            & Ours-w/o-Earbuds  & $10.709$  & $6.164$ & $10.247$ & $13.615$  & $14.642$ \\
            \cmidrule(l){2-7}
            & \multicolumn{6}{>{\columncolor{lightgray!50}}l}{\textbf{(b) Ablation on Cameras}} \\
            \cmidrule(l){2-7}
            & Ours-w/o-Cams & $11.284$ & $6.927$ & $11.219$ & $14.405$ & $14.441$\\
            & Ours-w/o-Cam\_f\_feature  & $9.691$ & $6.249$ & $10.080$ & $12.231$ & $13.211$ \\
            & Ours-w/o-Cam\_f\_SLAM  & $9.024$ & $5.966$ & $9.680$ & $11.299$ & $16.937$ \\
            \cmidrule(l){2-7}
            & \multicolumn{6}{>{\columncolor{lightgray!50}}l}{\textbf{(c) Ablation on Teacher-Student Architecture}} \\
            \cmidrule(l){2-7}
            & Ours-w/o-Teacher & $9.193$ & $6.258$ & $9.896$ & $11.631$ & $13.055$\\
            \midrule
            & \textbf{Ours} & $\textbf{8.459}$ & $\textbf{5.482}$ & $\textbf{9.035}$ & $\textbf{10.627}$ & $\textbf{12.382}$ \\
            \bottomrule
        \end{tabular}
    }%
\end{table*}

    
  

\subsection{Implementation and Metrics}
Full implementation details are provided in \supp.
To evaluate the proposed framework, we segment the dataset into non-overlapping sequences with a window size $N$, then randomly divide them into three subsets: $712$ sequences for training ($80\%$), $89$ for validation ($10\%$), and $89$ for testing ($10\%$).

Following prior work~\citep{mollyn2023imuposer,xu2024mobileposer,lee2024mocap}, we evaluate our model using the following metrics:
\textit{\acf{mpjpe}}: Mean Euclidean distance between predicted and ground-truth joint positions~(cm) with the pelvis aligned;
\textit{\acf{pa-mpjpe}}: \acs{mpjpe} after Procrustes alignment, measuring pose shape accuracy~(cm);
\textit{\acf{mpjre}}: Mean angular error between predicted and ground-truth joint orientations~(degrees) with the pelvis aligned;
\textit{\acf{mpjve}}: Mean per-vertex error across all vertices of the SMPL mesh~(cm) with the pelvis aligned;
\textit{\acf{root-pe}}: Root position error in global coordinates~(cm).

    
    
    
    

\subsection{Comparison}
To demonstrate the effectiveness of our approach, we conduct a comprehensive  comparison against two representative categories of baseline methods on our dataset: 

\textit{IMU-Only}\quad We compare against IMUPoser~\cite{mollyn2023imuposer} and MobilePoser~\cite{xu2024mobileposer}, sparse IMU methods using smartphones, smartwatches, and earbuds.

\textit{Camera-Only}\quad We adapt Fish2Mesh~\citep{shen2025fish2mesh} for two downward-facing cameras instead of one fisheye camera.

\begin{figure*}[t!]
    \centering
    \includegraphics[width=0.98\linewidth]{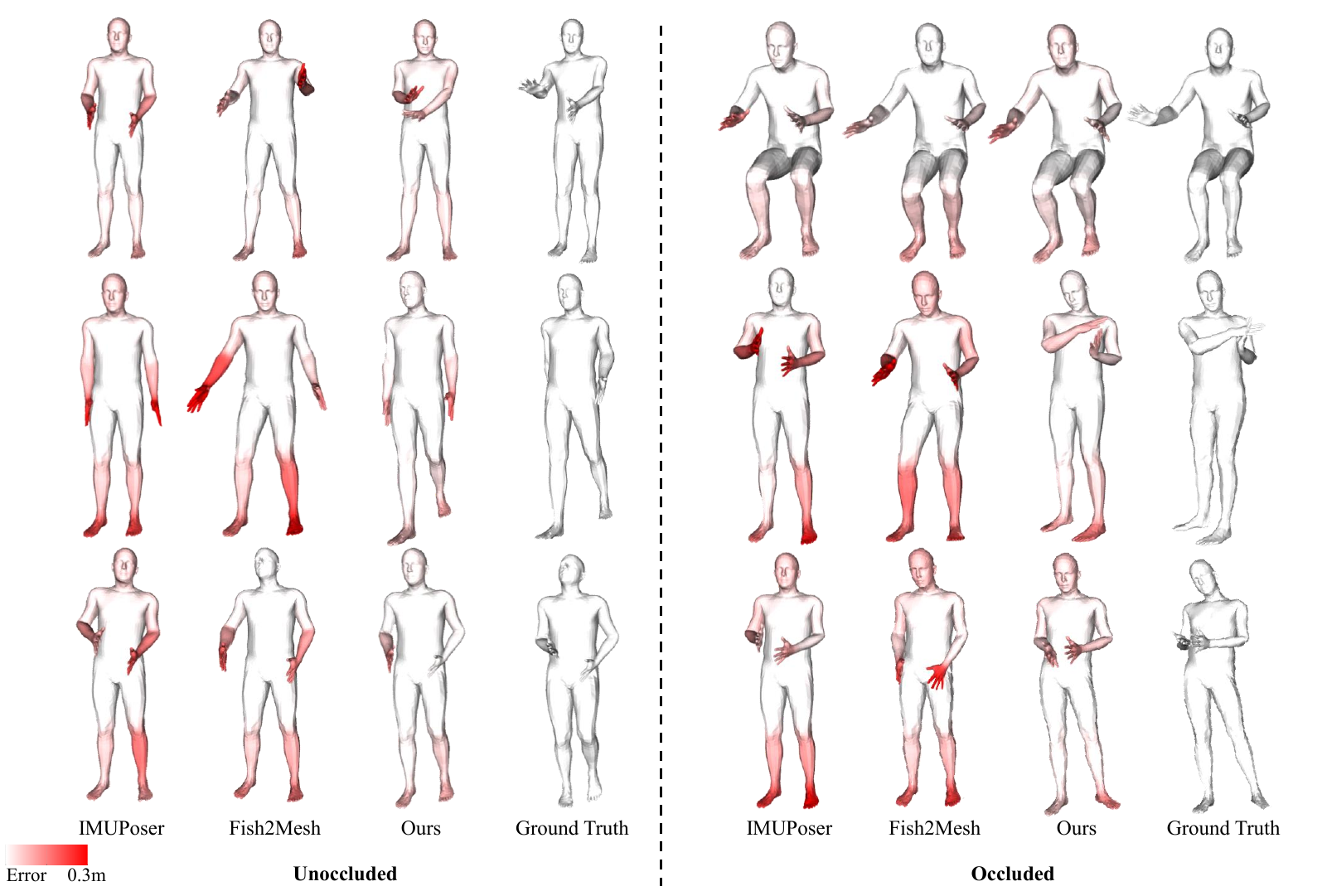}
    \caption{\textbf{Qualitative comparison.} This figure presents a comparison against baseline methods on Ego-Elec, where IMUPoser serves as the IMU-only baseline and Fish2Mesh represents the camera-only baseline. We visualize per-vertex SMPL error using a color map ranging from $0$ to $0.3$ m (white: low error, red: high error). Our method achieves consistently lower errors across diverse activities and maintains robustness under challenging occlusion scenarios.}
    \label{fig:results}
\end{figure*}

In \Cref{table:comparison}, we report quantitative results, which demonstrate that our full model outperforms all baselines across all metrics. The qualitative results in \cref{fig:results} also confirm these findings.  We also present results on challenging scenarios (sitting with occlusions, waving with body parts exiting the camera view), demonstrating our model's robustness across diverse conditions.

\paragraph{Comparison with IMU-based Baselines} 
We compare our full model against IMU-only baselines and our IMU-only ablation on our test set.
As shown in the first section of \cref{table:comparison}, our approach \textbf{significantly outperforms all baselines across all metrics}. 
Compared to the best baseline IMUPoser, our method achieves improvements of $3.35$ cm \ac{mpjpe}, $2.17$ cm \ac{pa-mpjpe}, $2.40$ deg \ac{mpjre}, $4.29$ cm \ac{mpjve}, and $12.12$ cm \ac{root-pe}, demonstrating the effectiveness of multimodal fusion.
Notably, our IMU-only variant also achieves improvement over IMUPoser despite using the same sensor configuration.
These \textbf{demonstrate three key insights}: (\rmnum{1}) the sparse $3$-IMU setup with loose everyday attachment poses significant challenges for pure inertial-based motion estimation. (\rmnum{2}) the addition of egocentric vision provides substantial benefits, enabling robust cross-modal compensation when individual sensors become unreliable due to noise or drift. (\rmnum{3}) our teacher-student distillation framework enables the student model to learn effective denoising and adaptation strategies for handling noisy and unstable consumer-grade IMU sensors.
    
    

\paragraph{Comparison with Camera-based Baselines} 
We compare Fish2Mesh~\citep{shen2025fish2mesh} against our full approach and our camera-only ablation. Our full model achieves $4.22$ cm and $5.59$ cm improvements over Fish2Mesh on \ac{mpjpe} and \ac{mpjve} respectively, with similar gains across other metrics. This improvement can be attributed to the fundamental limitations of camera-only Methods: body parts leaving the field of view, environmental and self-occlusions, and ambiguity in head pose. Our approach addresses these by \textbf{fusing visual observations with IMU signals, which provide motion and orientation cues when cameras fail}. Importantly, our camera-only variant performs comparably to Fish2Mesh, demonstrating that both vision-only approaches share similar limitations.


\begin{table}[t]
    \centering
    \caption{Comparison of our methods with baseline models on a typical challenging scenario, including occluded (sitting) and the body exiting the camera view (waving).}
    \label{tab:occluded}
    \resizebox{\linewidth}{!}{%
        \begin{tabular}{lcccc}%
            \toprule
            \multirow{2.4}{*}{Method} & \multicolumn{2}{c}{Sitting} & \multicolumn{2}{c}{Waving}\\
            \cmidrule(lr){2-3} \cmidrule(lr){4-5}   & Upper PE & Lower PE  & Upper PE & Lower PE\\
            \midrule
            IMUPoser \cite{mollyn2023imuposer}  & $9.230$ & $8.946$ & $21.197$ & $12.472$ \\
            Fish2Mesh \cite{shen2025fish2mesh}  & $9.369$  & $6.058$ & $22.582$ & $12.052$\\
            
            \midrule
            \textbf{Ours}  & $\textbf{5.315}$ & $\textbf{4.541}$  & $\textbf{18.465}$ & $\textbf{9.674}$\\
                
            \bottomrule
        \end{tabular}%
    }%
\end{table}

\paragraph{Comparison in Occluded Scenario}
\Cref{tab:occluded} evaluates performance on challenging scenarios: sitting (lower body occlusion) and waving (upper body exits view). We report region-specific Upper-\ac{mpjpe} and Lower-\ac{mpjpe}, which measure upper-body and lower-body joint errors, respectively.
While baseline methods struggle with these challenging scenarios, achieving comparable errors to each other, our approach significantly outperforms them by approximately $4$ cm on both upper and lower body metrics. This demonstrates effective multimodal fusion: IMU sensors compensate when cameras are occluded, while cameras track visible regions when body parts leave the frame.

\paragraph{Qualitative Comparisons}
\Cref{fig:results} presents qualitative comparisons across unoccluded 
and occluded scenarios.
For the unoccluded scenes, rows $1$-$2$ show camera-only methods failing when the upper body exits the field of view, while our method maintains accuracy via IMU data. Rows $2$-$3$ demonstrate that IMU-only methods struggle to estimate leg movements with sparse sensors, a challenge our approach addresses by incorporating complementary visual information.
For the occluded scenes, the last rows show scenarios where objects (boxes, books) block camera views. Camera-only methods fail under these occlusions, whereas 
our multimodal fusion enables robust tracking by leveraging IMU signals. This \textbf{cross-modal compensation and teacher-student distillation are key to handling real-world challenges}.

\begin{figure}[t!]
    \centering
    \includegraphics[width=\linewidth]{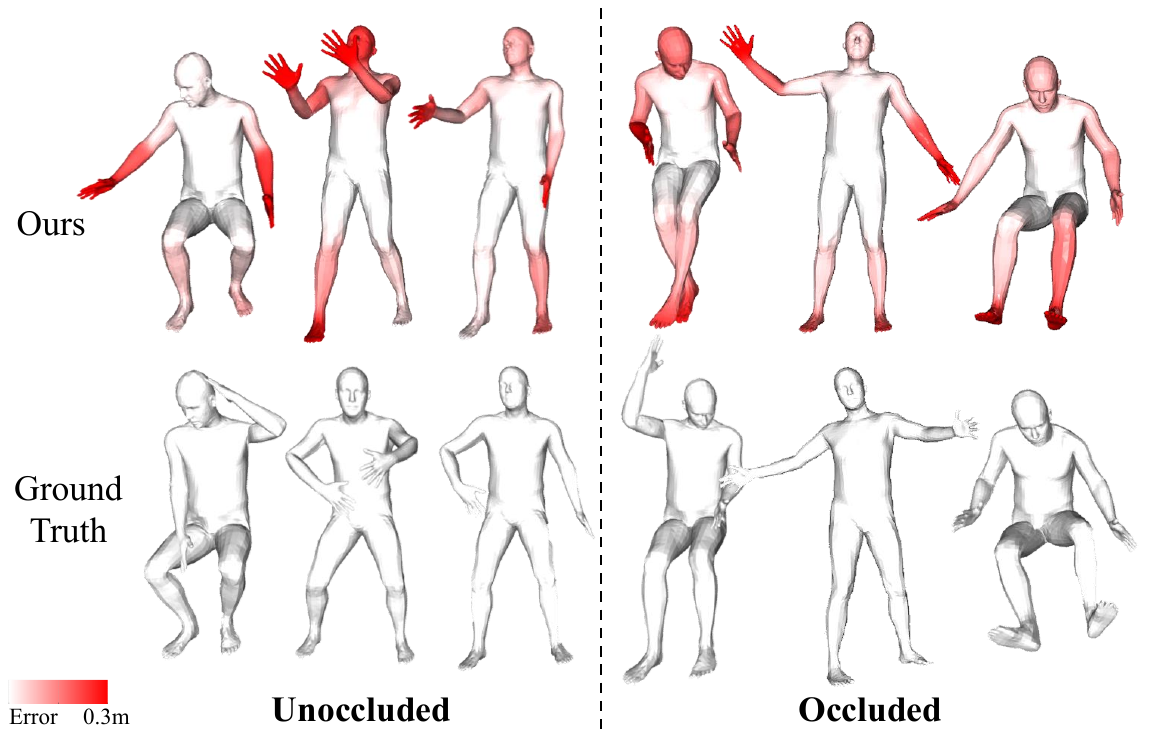}
    \caption{\textbf{Failure cases.} We show rare failure scenarios where loose attachment of consumer devices (phone shifting in pocket, watch with play) introduces sensor noise and instability. }
    \label{fig:failure}
\end{figure}

\paragraph{Failure Cases} \Cref{fig:failure} shows our method struggles when consumer IMU sensors become unstable due to loose attachment or sensor noise. Our model fails when the upper body exits the camera view while IMU signals are simultaneously unreliable, though the lower body maintains reasonable tracking. In severely occluded scenarios, our model exhibits reduced accuracy for occluded body parts while maintaining overall motion structure. 

\subsection{Ablation study}
We conducted systematic ablations to assess the impact of various model configurations on human motion estimation performance, as shown in the second section of \cref{table:comparison}.

\paragraph{Abalation on IMUs}

We ablate IMU sensor configurations, and the results show that removing all IMUs degrades performance but remains comparable to the camera-only baseline, validating our model's adaptability.
Individual sensor ablations reveal that earbuds contribute most: removing earbuds causes $2.25$ cm \ac{mpjpe} and $2.99$ cm \ac{mpjve} degradation versus $1$-$2$ cm for others. Notably, \ac{mpjre}, which measures the local pose, degrades minimally while \ac{mpjpe} and \ac{mpjve} increase  $>2$ cm, indicating earbuds' critical role in global orientation estimation.

\paragraph{Ablation on Cameras}
We conduct camera ablation studies with three configurations: removing all 
cameras (w/o-Cams), only excluding the forward-facing camera from motion estimation~(w/o-Cam\_f\_feature), or only excluding it from SLAM~(w/o-Cam\_f\_SLAM).

\textit{Individual Camera Contributions}\quad Results show expected patterns based on camera viewpoints. Removing the forward-facing camera from SLAM degrades \ac{root-pe} (global localization) most severely, increasing error by $4.56$ cm. This is expected, as SLAM provides environmental context critical for global positioning. Conversely, removing the forward-facing camera from motion estimation degrades \ac{mpjpe} by $1.23$ cm, as this camera directly observes the upper body.

\textit{All Cameras Removed (w/o-Cams)}\quad When all cameras are removed (IMU-only configuration), \ac{mpjpe} increases by $2.83$ cm and \ac{mpjve} by $3.78$ cm, demonstrating substantial degradation. This validates the importance of multimodal fusion: visual information significantly outperforms IMU-only estimation.

\textit{Interesting Finding}\quad We observe a counterintuitive result: removing \textit{all} cameras (\ac{root-pe} = $14.4$ cm) achieves better global localization than removing only the forward camera (\ac{root-pe} = $16.9$ cm). This suggests that downward-facing cameras, while beneficial for body pose estimation, may introduce conflicting signals for global localization when environmental context from the forward camera is absent. The model handles pure IMU-based global tracking more robustly than partial visual information lacking environmental context.

\paragraph{Teacher-Student Model}
We ablate the teacher-student framework by training without distillation. Performance degrades by $0.8$ cm across all metrics (\ac{mpjpe}, \ac{pa-mpjpe}, \ac{mpjve}), confirming that the distillation enables effective adaptation to noisy consumer sensors.

\section{Discussions}

\paragraph{Conclusions}
We present Ego-Elec, a lightweight human motion capture method using everyday wearables without calibration, enabled by multimodal fusion and teacher-student distillation. Through this distillation paradigm, our model learns to adapt to noisy real-world consumer sensors. We also introduce a large-scale real-world dataset spanning diverse daily activities and environments.
By achieving state-of-the-art performance and providing comprehensive real-world data, our work establishes a foundation for future research on full-body motion estimation with consumer devices, with applications in XR gaming, telepresence, and humanoid teleoperation.

\paragraph{Limitations}

Our current approach operates offline, but can be adapted to real-time use for widespread real-world applications. The IMU configuration currently supports only fixed sensor placements and requires prior specification, which can be improved by enabling adaptive handling of variable sensor configurations. Future work will further incorporate richer interaction modalities to enhance robustness in complex real-world environments.

\clearpage
{
    \small
    \bibliographystyle{ieeenat_fullname}
    \bibliography{reference_header, main}
}

\clearpage
\appendix
\renewcommand\thefigure{A\arabic{figure}}
\setcounter{figure}{0}
\renewcommand\thetable{A\arabic{table}}
\setcounter{table}{0}
\renewcommand\theequation{A\arabic{equation}}
\setcounter{equation}{0}
\pagenumbering{arabic}%
\renewcommand*{\thepage}{A\arabic{page}}
\setcounter{footnote}{0}
\setlength\floatsep{1\baselineskip plus 3pt minus 2pt}
\setlength\textfloatsep{1\baselineskip plus 3pt minus 2pt}
\setlength\dbltextfloatsep{1\baselineskip plus 3pt minus 2 pt}
\setlength\intextsep{1\baselineskip plus 3pt minus 2 pt}
\maketitlesupplementary

\section{Ego-Elec Dataset} \label{supp:sec:dataset}

We provide statistical analysis of the \dataset dataset, characterizing environment types, activity categories, activity durations, and their distributions.
\begin{figure}[ht]
    \centering
    \includegraphics[width=\linewidth]{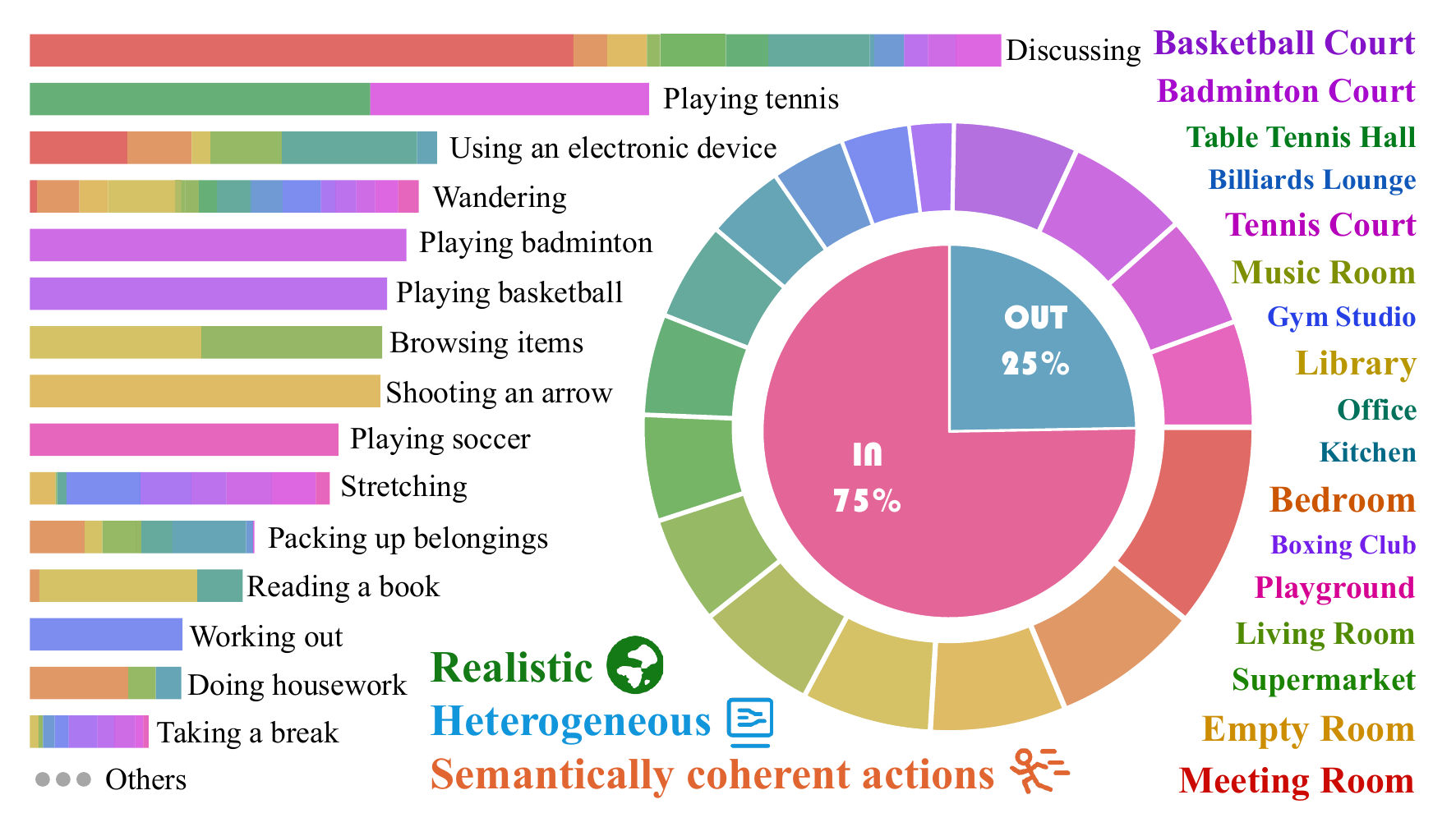}
    \caption{\textbf{Pie Chart: Distribution of Environments.} The dataset comprises a variety of environments, with their respective quantities illustrated by the proportions shown in the pie chart. \textbf{Bar Chart: Distribution of Activity Types.} \dataset includes diverse activity catgories. Here, we present the most frequent activities, along with the distribution of environments associated with each activity.}
    \label{sup:fig:datasets}
\end{figure}
\paragraph{Environment Diversity} As shown in the pie chart in \cref{sup:fig:datasets}, our dataset spans $17$ distinct environment types across indoor and outdoor settings, with their relative proportions indicated. This diversity reflects the variety of real-world scenarios encountered in daily life.

\begin{figure}[t!]
    \centering
    \includegraphics[width=\linewidth]{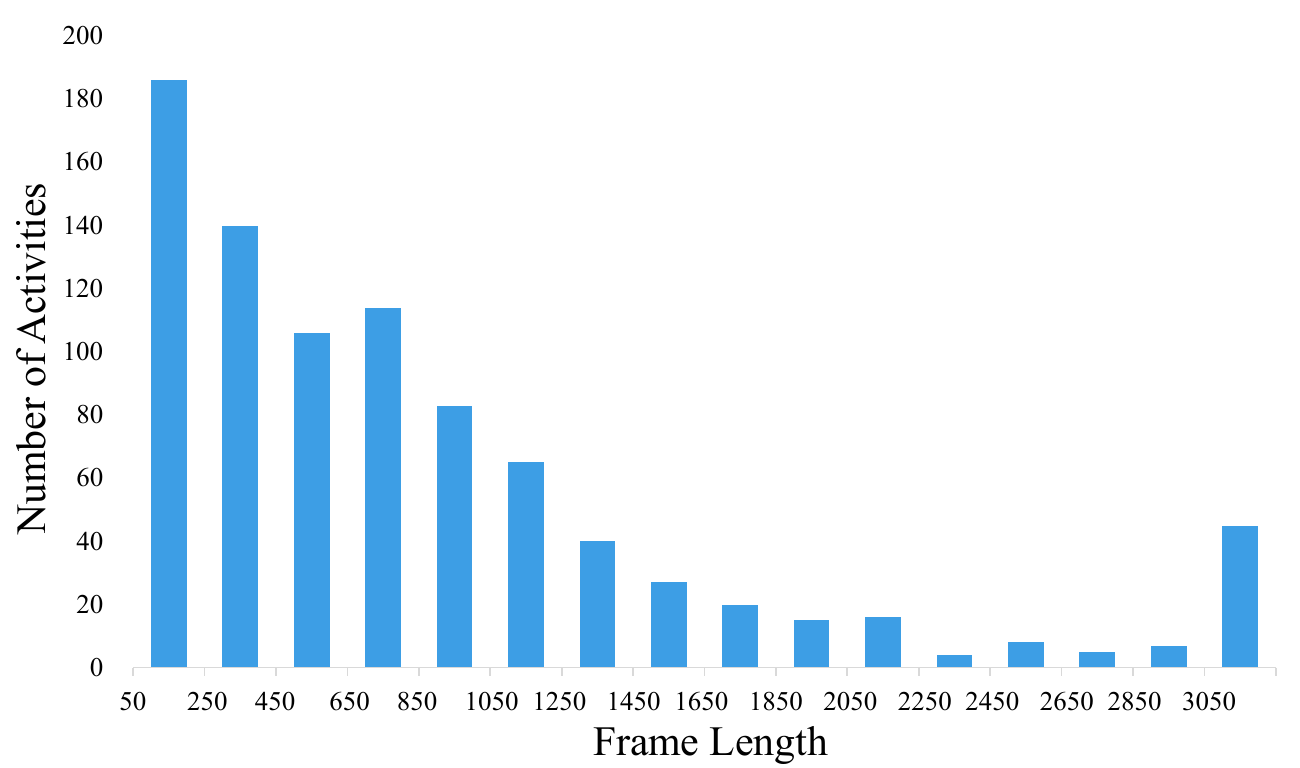}
    \caption{\textbf{Frame Length Distribution across Activities.} This figure shows the distribution of frame lengths across different activities, ranging from about $50$ frames to several thousand.}
    \label{sup:fig:group}
\end{figure}

\paragraph{Activity Coverage} The bar chart in \cref{sup:fig:datasets} shows the $56$ most common daily activities in our dataset. Activity durations range from $50$ to several thousand frames (\cref{sup:fig:group}), capturing both short actions and extended activities.

\paragraph{Data Collection Protocol} We recorded full sessions containing $5$–$10$ activities arranged in contextually meaningful sequences, then segmented each session into individual activities.

\section{System Setup}


\begin{figure}
    \centering
    \includegraphics[width=0.95\linewidth]{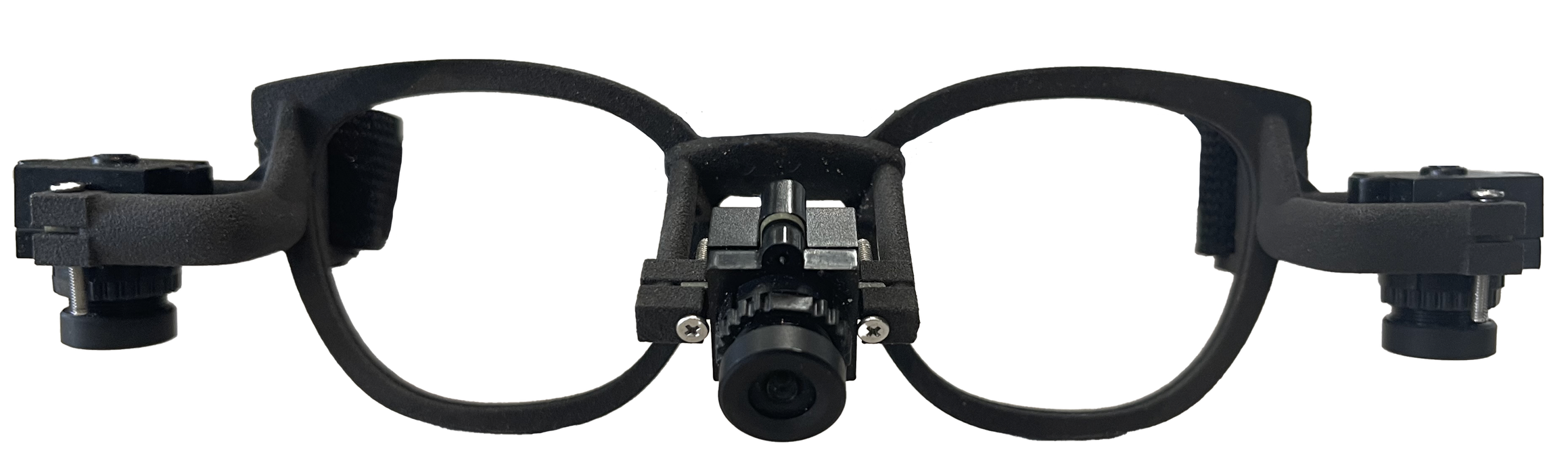}
    \caption{\textbf{Glasses Prototype.} The glasses prototype is equipped with one forward-facing camera and two downward-facing cameras, and it runs on a Strawberry $5$ platform.}
    \label{supp:fig:prototype}
\end{figure}

\subsection{Glasses Prototype}
We developed a glasses prototype (\cref{supp:fig:prototype}) with three RGB cameras ($1280\times 720$ @ $30$ FPS, $120^{\circ}$ FOV): one forward-facing camera for environmental observation, and two downward-facing cameras (left and right) for egocentric body observation.

We will release all hardware designs, assembly instructions, calibration 
procedures, and data collection code to facilitate future research.


\subsubsection{Camera Synchronization}

To achieve frame-level synchronization across the three cameras, we implement a software-based synchronization mechanism built around three core components: a \textbf{barrier-based capture trigger}, a \textbf{global frame index system}, and a \textbf{asynchronous storage pipeline}. Each camera operates in its own capture thread, coordinated by a centralized controller that manages shared synchronization primitives.

\paragraph{Barrier-Based Capture Trigger}
Each camera thread runs continuously but must wait at a shared barrier before capturing the next frame. The barrier releases only when all three threads have reached it, triggering simultaneous frame capture. This mechanism eliminates temporal drift caused by asynchronous execution or uneven driver latency.

\paragraph{Global Frame Index System}
A global frame counter ensures consistent labeling across all camera views. In each synchronized capture cycle, the first thread that exits the barrier assigns the next global frame index, and the remaining threads reuse that same index. Immediately after capture, each thread records a timestamp and pushes it into a camera-specific queue. This guarantees temporally consistent timestamps across all cameras.

\paragraph{Asynchronous Storage Pipeline}
To prevent disk I/O latency from interrupting capture operations, all frames are placed into per-camera queues and saved asynchronously by dedicated writer threads. Timestamps are flushed only after all cameras have submitted the same global frame index, ensuring completeness. This design maintains high capture throughput and prevents queue backpressure from degrading synchronization accuracy.




\section{Implementation Details} \label{supp:sec:impl}

\subsection{Data Alignment}
Since the cameras, everyday wearables, and \ac{mocap} system run on different devices, their timestamps are not inherently synchronized. To achieve temporal alignment across all modalities, we introduce two simple alignment gestures performed at the start of each recording session.

\textbf{Gesture 1 (Wearables–\ac{mocap} Alignment)}
The user extends their right arm with the palm facing upward, places the smartphone on the palm, and raises the arm vertically three times while keeping the phone horizontal. Using this gesture, we compute the z-axis acceleration from both the smartphone IMU and the \ac{mocap} right-hand sensor. By temporally aligning the peak responses from the three repeated arm raises, we compute the time offset needed to synchronize the smartphone IMU with the \ac{mocap} system. Since the smartphone, smartwatch, and earbuds share a synchronized clock (all connected to the same mobile device), aligning the smartphone with the \ac{mocap} system automatically synchronizes all consumer wearable devices.

\textbf{Gesture 2 (Cameras–\ac{mocap} Alignment)}
The user stands upright, quickly turns their head to the right, and then returns to a forward-facing position. From the camera streams, we select the frame where the camera's field of view reaches its farthest rightward extent, corresponding to the moment of maximum head rotation. We then match this frame to the extremum of the z-axis orientation recorded by the \ac{mocap} head sensor, yielding the time offset between the cameras and the \ac{mocap} system for alignment.

Together, these two gestures provide accurate temporal synchronization across the cameras, wearables, and the \ac{mocap} ground-truth system.

\subsection{Data Preprocess}
Input images are resized to $224 \times 224$ for visual feature encoding, while the SLAM module operates on the original camera resolution ($1280\times 720$) to retain sufficient spatial detail for reliable tracking.
To reduce high-frequency noise in the IMU signals while preserving natural motion dynamics, we apply a sliding-window smoothing filter with a window size of $5$.
All ground-truth global translations are expressed in a head-relative coordinate frame, computed as the relative transformation with respect to the first frame of each sequence.


\subsection{Network Architecture}
The teacher and student models share the same network architecture. The visual feature encoder is implemented using a ResNet backbone pre-trained on ImageNet, producing a $512$-dimensional feature vector. The IMU feature encoder is an MLP with layer sizes ($60$, $256$, $128$). The resulting latent features are then concatenated with the head pose estimated from the SLAM module. Finally, the fused representation is passed through a bidirectional LSTM, which outputs the predicted motion representation $\mathcal{M}$.

\begin{table}[t]
    \centering
    \caption{\textbf{Per-joint weights used in $\mathcal{L}_{pose}$.}}
    \label{sup:tab:joint_weights}
        \begin{tabular}{l c c c c}
            \toprule
            \textbf{Joint Name} & \textbf{Weight} & \textbf{Joint Name} & \textbf{Weight} \\
            \midrule
            Pelvis & $1.0$ & Neck & $0.1$ \\
            Left Hip & $0.2$ & Left Collar & $0.3$ \\
            Right Hip & $0.2$ & Right Collar & $0.3$ \\
            Spine1 & $0.1$ & Head & $0.3$ \\
            Left Knee & $0.3$ & Left Shoulder & $0.2$ \\
            Right Knee & $0.3$ & Right Shoulder & $0.2$ \\
            Spine2 & $0.1$ & Left Elbow & $0.3$ \\
            Left Ankle & $0.3$ & Right Elbow & $0.3$ \\
            Right Ankle & $0.3$ & Left Wrist & $0.4$ \\
            Spine3 & $0.1$ & Right Wrist & $0.4$ \\
            Left Foot & $0.3$ & Left Hand & $0.4$ \\
            Right Foot & $0.3$ & Right Hand & $0.4$ \\
            \bottomrule
        \end{tabular}
\end{table}




\subsection{Training Details}
We train both teacher and student models for $500$ epochs with batch size $256$ on an NVIDIA RTX $4080$ GPU using the Adam optimizer with window size $N=50$ frames.

\textbf{Teacher Training} We optimize $\mathcal{L}_{tea}$ with learning rate $1 \times 10^{-3}$. For $\mathcal{L}_{pose}$, we employ per-joint weights $w_j$ (\cref{loss:local_pose}) to emphasize challenging joints: the global pelvis rotation and limb joint rotations, which are often difficult to estimate accurately from egocentric observations, the loss weights are shown in~\cref{sup:tab:joint_weights}.

\textbf{Student Training} We initialize the student with the teacher's pre-trained weights, except for the student IMU encoder, which is randomly initialized. The MLP-based IMU encoder is trained with a learning rate $1 \times 10^{-3}$ to learn representations that compensate for missing dense IMU observations. The remaining network parameters are fine-tuned at a lower learning rate of $1 \times 10^{-4}$ to preserve their learned capacity for mapping multi-modal features to motion predictions.

\textbf{Student Loss Weighting} We initialize the loss coefficients in $\mathcal{L}_{stu}$ (\cref{loss:stu}) as: $\lambda_{motion} = 1.0$, $\lambda_{output} = 0.5$, and $\lambda_{feat} = 0.5$. The latter two are decayed by a factor of $0.8$ every $10$ epochs to gradually shift emphasis toward direct motion prediction.

\end{document}